\documentclass{article}

\usepackage{PRIMEarxiv}

\usepackage[utf8]{inputenc} 
\usepackage[T1]{fontenc}    
\usepackage{hyperref}       
\usepackage{url}            
\usepackage{booktabs}       
\usepackage{amsfonts}       
\usepackage{nicefrac}       
\usepackage{microtype}      
\usepackage{lipsum}
\usepackage{placeins}
\usepackage{fancyhdr}       
\usepackage{graphicx}       
\graphicspath{{media/}}     
\usepackage{orcidlink}
\usepackage{xcolor}
\usepackage{tikz}
\usepackage{adjustbox}
\usepackage{subcaption}
\usepackage{booktabs}
\usepackage{multirow}
\usepackage{todonotes}

\usepackage{glossaries-extra}
\setkeys{glslink}{hyper=false}
\newabbreviation{ad}{AD}{anomaly detection}
\newabbreviation{admd}{ADMD}{Anomaly Detection Meta-data}
\newabbreviation{ae}{AE}{Autoencoder}
\newabbreviation{auc}{AUC}{Area Under the Curve}

\newabbreviation{bilstm}{BiLSTM}{bidirectional Long Short-Term Memory}

\newabbreviation{cnn}{CNN}{Convolutional Neural Network}
\newabbreviation{cnnbilstm}{CNN-BiLSTM}{Convolutional and Bidirectional Long Short-Term Memory Network}
\newabbreviation{cps}{CPS}{Cyber Physical System}
\newabbreviation{csv}{CSV}{Comma-Separated Values}

\newabbreviation{dbm}{DBM}{deep Boltzmann machine}
\newabbreviation{dbn}{DBN}{deep belief network}
\newabbreviation{dnn}{DNN}{deep neural network}
\newabbreviation{dl}{DL}{Deep Learning}
\newabbreviation{dt}{DT}{Decision Tree}
\newabbreviation{dvc}{DVC}{Data Version Control}

\newabbreviation{gfl}{GFL}{Graph Feature Learning}
\newabbreviation{gru}{GRU}{Gated Recurrent Unit}

\newabbreviation{ics}{ICS}{Industrial Control System}
\newabbreviation{ids}{IDS}{Intrusion Detection Systems}
\newabbreviation{iot}{IoT}{Internet of Things}
\newabbreviation{ip}{IP}{Internet Protocol}
\newabbreviation{ipfix}{IPFIX}{Internet Protocol Flow Information Export}

\newabbreviation{kl}{KL}{Kullback-Leibler}

\newabbreviation{lr}{LR}{Logistic Regression}
\newabbreviation{lstm}{LSTM}{Long Short-Term Memory}

\newabbreviation{mawi}{MAWI}{Measurement and Analysis on the WIDE Internet}
\newabbreviation{ml}{ML}{Machine Learning}
\newabbreviation{mlp}{MLP}{Multi-Layer Perceptron}

\newabbreviation{pca}{PCA}{Principal Component Analysis}
\newabbreviation{pcap}{pcap}{packet capture}

\newabbreviation{rf}{RF}{Random Forest}
\newabbreviation{rnn}{RNN}{Recurrent Neural Network}
\newabbreviation{roc}{ROC}{Receiver Operating Characteristic}

\newabbreviation{silk}{SiLK}{System for Internet-Level Knowledge}
\newabbreviation{smote}{SMOTE}{Synthetic Minority Over-sampling Technique}
\newabbreviation{snmp}{SNMP}{Simple Network Management Protocol}

\newabbreviation{tcp}{TCP}{Transmission Control Protocol}

\newabbreviation{xml}{XML}{Extensible Markup Language}
\GlsXtrEnableEntryCounting
 {abbreviation}
 {1}

\definecolor{plot1}{HTML}{1f77b4}
\definecolor{plot2}{HTML}{ff7f0e}
\definecolor{plot3}{HTML}{2ca02c}
\definecolor{plot4}{HTML}{d62728}

\newcommand{\linestylebox}[2]{%
  \begin{tikzpicture}[baseline=-0.5ex]
    \draw[#1, ultra thick, #2] (0,0) -- (0.75,0);
  \end{tikzpicture}%
}

\title{MAWIFlow Benchmark: Realistic Flow-Based Evaluation for Network Intrusion Detection
}

\author{
  Joshua Schraven\orcidlink{0009-0003-8245-9228},
  Alexander Windmann\orcidlink{0000-0002-6522-4262},
  Oliver Niggemann\orcidlink{0000-0001-8747-3596} \\
  Institute of Artificial Intelligence \\
  Helmut Schmidt University \\
  Hamburg, Germany\\
  \texttt{\{joshua.schraven,alexander.windmann,oliver.niggemann\}@hsu-hh.de} \\
}

\begin{document}
\maketitle

\begin{abstract}
Benchmark datasets for network intrusion detection commonly rely on synthetically generated traffic, which fails to reflect the statistical variability and temporal drift encountered in operational environments.
This paper introduces MAWIFlow, a flow-based benchmark derived from the MAWILAB~v1.1 dataset, designed to enable realistic and reproducible evaluation of anomaly detection methods.
A reproducible preprocessing pipeline is presented that transforms raw packet captures into flow representations conforming to the CICFlowMeter format, while preserving MAWILab’s original anomaly labels.
The resulting datasets comprise temporally distinct samples from January 2011, 2016, and 2021, drawn from trans-Pacific backbone traffic.

To establish reference baselines, traditional machine learning methods, including Decision Trees, Random Forests, XGBoost, and Logistic Regression, are compared to a deep learning model based on a CNN-BiLSTM architecture.
Empirical results demonstrate that tree-based classifiers perform well on temporally static data but experience significant performance degradation over time.
In contrast, the CNN-BiLSTM model maintains better performance, thus showing improved generalization.
These findings underscore the limitations of synthetic benchmarks and static models, and motivate the adoption of realistic datasets with explicit temporal structure.
All datasets, pipeline code, and model implementations are made publicly available to foster transparency and reproducibility\footnote{\url{https://github.com/TheLurps/MAWIFlow}}.
\end{abstract}

\keywords{Network Intrusion Detection \and Anomaly Detection Benchmark \and Temporal Drift \and MAWILab}

\section{Introduction}\label{sec:introduction}

Network \gls{ad} represents a critical component in modern cybersecurity strategies, integral to safeguarding infrastructures from increasingly sophisticated cyber threats.
The continuous growth and diversification of network-based attacks highlight the necessity for robust and adaptive \gls{ad} techniques capable of promptly identifying and mitigating malicious activities.

Conventional approaches to network security, particularly signature-based \gls{ids}, rely predominantly on predefined patterns or signatures representing known threats.
Although highly effective against previously cataloged attacks, these approaches suffer fundamental limitations.
Firstly, continual updates are mandatory to recognize emerging threats, making their maintenance labor-intensive and costly.
Secondly, signature-based methods do not scale efficiently with the exponential volume and complexity of traffic in contemporary networks; thus, their practicability diminishes progressively over time as network traffic continues to expand rapidly.

To address these shortcomings, research has increasingly pivoted towards \gls{ml} and \gls{dl} methods, able to recognize anomalies by learning data-driven patterns from observed network behaviors rather than manually curated signatures.
However, the efficacy of such advanced methods in practical, large-scale network operational environments often diverges significantly from laboratory benchmark performance, highlighting an ongoing gap between controlled evaluations and realistic deployment scenarios.
While numerous widely utilized benchmark datasets exist, such as CIC-IDS-2017 and CSE-CIC-IDS2018~\cite{sharafaldin2018generating}, these predominantly comprise synthetically generated network traffic designed to approximate realistic user activities.
Though valuable for algorithmic development, synthetic datasets inherently fall short of encapsulating the full complexity and variability observed in real-world network traffic contexts.
Specifically, they often fail to represent the continuous evolution of threat landscapes, nuanced traffic behaviors, and realistic anomaly distributions, calling into question their usefulness for comprehensive method evaluations. 

To support realistic and reproducible assessments of network \gls{ad} methods, this work contributes the following core elements:

\begin{enumerate}
\item A structured, reproducible, and open-source preprocessing pipeline for transforming raw MAWILab~v1.1 datasets~\cite{fontugne2010mawilab} into standardized flow-based representations compatible with the widely adopted CICFlowMeter~\cite{draper-gil_characterization_2016,habibi_lashkari_characterization_2017} format. 
\item A corresponding novel benchmark dataset derived from MAWILab~v1.1 via the aforementioned pipeline. 
\item A rigorous comparative evaluation framework, benchmarking the efficacy of traditional machine learning algorithms and a selected deep learning model across multiple temporal windows within MAWILab, contrasted against conventional synthetic benchmarks (CIC-IDS-2017).
\end{enumerate}

The remainder of the paper is structured as follows: Section~\ref{sec:related-work} reviews related work on network anomaly detection and dataset design.
Section~\ref{sec:proposed-benchmark} details the MAWILab~v1.1 dataset, its labeling process, and our reproducible \gls{dvc}-powered pipeline for converting raw captures into CICFlowMeter-compatible flows as well as introducing our evaluation strategy.
Section~\ref{sec:evaluation} presents experimental results: within-dataset baselines on CIC-IDS-2017 and MAWIFlow (2011, 2016, 2021) and cross-temporal generalization analyses.
Section~\ref{sec:discussion} discusses the implications of our findings, highlighting the limitations of synthetic benchmarks, the impact of temporal drift, and strategies for robust anomaly detection.
Finally, Section~\ref{sec:conclusion} concludes and suggests future directions.

\section{Related Work}\label{sec:related-work}

\gls{ad} on network data has been achieved using several different methods.
They can be categorized into the following groups as shown by \cite{zhang2024survey}: statistical, \gls{ml}, \gls{dl}, behavior analysis, and hybrid methods.
From the traditional \gls{ml} based approaches, \gls{dt} based methods such as \glspl{rf}~\cite{abdelaziz2024EnhancingNetworkThreat} and XGBoost~\cite{iftikhar2025IntrusionDetectionNSLKDD} as well as \gls{lr}~\cite{kolukisa2024EfficientNetworkIntrusion} have proven to be effective.
Looking at the available \gls{dl} methods, the usage of \glspl{cnn}~\cite{deshmukh2024EfficientCNNBasedIntrusion}, \glspl{rnn} like \gls{lstm}~\cite{dash2025OptimizedLSTMbasedDeep} and \glspl{gru}~\cite{shoab2024GRUEnabledIntrusion}, Autoencoders~\cite{xu2021ImprovingPerformanceAutoencoderBased}, Transformers~\cite{manocchio2024FlowTransformerTransformerFramework}, and various combinations can be observed.
A promising model is described and evaluated by \cite{wang2024new} using a \gls{bilstm} on the CICIDS-2017 dataset~\cite{sharafaldin2018generating} and a gas pipeline dataset obtained by a research team at Mississippi State University~\cite{turnipseed2015new}.
The proposed model outperformed models using \glspl{mlp}~\cite{belarbi2022intrusion,roopak2019deep}, \gls{lstm}~\cite{roopak2019deep} and Deep\gls{gfl}~\cite{yao2018deepgfl}.

Beyond detection techniques, the characteristics of evaluation datasets  themselves have a major impact on \gls{ids} research. Datasets can be categorized according to their form of data and their scenario.
As described by \cite{fernandes2019comprehensive}, network data can be categorized into the following types: \gls{tcp} dump, \gls{snmp} and \gls{ip} Flow.
\gls{tcp} dumps are often created by using port mirroring, stored as \glspl{pcap} and contain recorded network packets.
Often, packet payloads are removed to sanitize the data.
Using \glspl{pcap} allows for the easiest and most detailed recording strategy.
\gls{snmp} is mostly telemetry data and recorded from edges of the network such as spine switches or routers.
These allow a much higher level of abstraction than \gls{tcp} dumps that use less space in storage.
\gls{ip} Flow formats such as NetFlow~\cite{claise2004rfc3954}, \glsxtrshort{ipfix}~\cite{claise2013rfc7011} and sFlow~\cite{phaal2001rfc3176} provide a high-level summary of network traffic.
They contain computed fields such as the number of packets, total byte size, flow direction, type of service, session count, etc.
Most of this data can be generated from \gls{pcap} recordings.
Regarding scenarios, one can differentiate between standard computer networks, \gls{ics} networks and \gls{iot} networks, where standard computer networks describe networks with one or multiple layers of interconnected devices that provide services to an end user.
\gls{ics} and \gls{iot} networks differ from these in terms of the protocols used and their lower complexity.
Commonly used benchmark datasets such as NSL-KDD~\cite{tavallaee2009detailed} or Kyoto 2006+~\cite{song2011statistical} consist of \gls{tcp} dumps from standard computer networks.
Common benchmark datasets utilizing the same scenario but containing \gls{ip} flows are UNSW-NB15~\cite{moustafa2015unswnb15} and CIC-IDS2017~\cite{sharafaldin2018generating}.

However, all of these datasets consist of synthetic data. Although datasets such as CIC-IDS2017 and its follow-up CSE-CIC-IDS2018~\cite{sharafaldin2018generating} attempt to mimic realistic user activities, they are inherently unable to capture current attack patterns.
Being somewhat dated, they fail to represent modern threat behaviors.
In addition, anomaly classes tend to be unevenly dispersed, and the assumptions made during simulations remain constant throughout the simulation period.
Notably, the majority of \gls{ids} datasets do not account for long-term changes in network traffic (e.g. datasets rarely span samples across multiple years).
This static nature prevents the assessment of changing attack strategies or user behavior, limiting the evaluation of the robustness of a model.
Therefore, a dataset consisting of real data such as MAWILab~v1.1~\cite{fontugne2010mawilab} is beneficial.
It consists of \gls{pcap} files recorded on a trans-Pacific link between Japan and the United States over 18 years.
Labels are added using an ensemble of different detection methods as described by \cite{fontugne2010mawilab}.
A first attempt to turn the raw MAWILab packet traces into labelled flow records was presented by \cite{kim2019GeneratingLabeledFlow}.
Their two-step pipeline employs the \gls{silk} to export NetFlow-v9 tuples and then attaches anomaly labels by matching source and destination \glspl{ip} as well as ports.
While this method demonstrated the general feasibility of re-using MAWILab for flow-based studies, it omits the start/stop time windows carried in the annotations and therefore risks assigning a label to traffic that merely reuses the same combination of \glspl{ip} and ports outside the alarm window.
In addition, the authors drop the \texttt{notice} class altogether and introduce an ad-hoc \texttt{unsure} category when only a single attribute matches.
These design choices simplify the mapping but reduce label fidelity---an issue that our MAWIFlow pipeline addresses by packet-level splitting before flow generation and by propagating the full MAWILab label semantics. Furthermore, \cite{kim2019GeneratingLabeledFlow} only presents a pipeline for one day of data and omits analysis and evaluation of the resulting flows.

\section{Proposed Benchmark}\label{sec:proposed-benchmark}

The proposed benchmark for \gls{ad} on network data builds upon the MAWILab dataset~\cite{fontugne2010mawilab}.
This paper proposes to extend MAWILab~v1.1 with a preprocessing pipeline that enables the detection of deteriorating performance of \gls{ad} methods over time.
We also train and test multiple classifiers to establish a baseline performance.

\subsection{MAWILab Dataset}\label{subsec:mawilab-dataset}

MAWILab~v1.1~\cite{fontugne2010mawilab}, is a real-world dataset designed to support the evaluation of network \gls{ad} methods.
It consists of daily 15-minute \gls{pcap} recordings from a trans-Pacific backbone link between Japan and the United States.
The dataset spans from 2007 to 2024 and contains compressed traces in both \gls{tcp} dump and \gls{pcap} formats, accompanied by anomaly annotations.
These annotations are provided in two formats per recording day: a \gls{csv} file listing filter expressions and an \gls{admd} file.
While both formats encode the IP and port filters associated with each detected anomaly, only the \gls{admd} format includes temporal criteria.
In total, the dataset is estimated to have a size of approximately 6.5 terabytes.

\begin{figure}[b]
    \centering
    \includegraphics[width=0.8\textwidth]{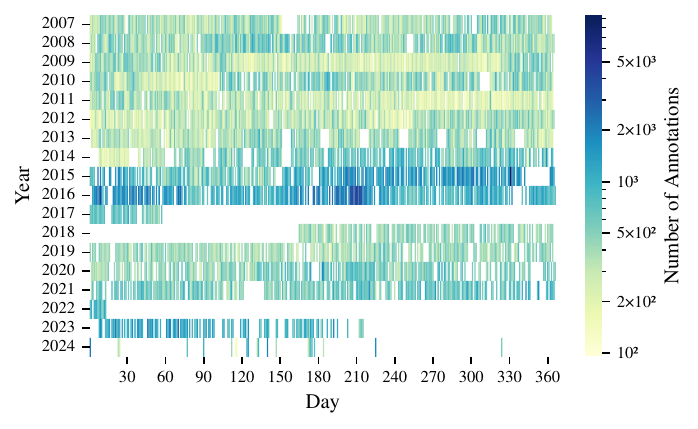}
    \caption{Daily count of anomaly annotation filters in MAWILab~v1.1 from 2007 to 2024, plotted on a logarithmic scale to highlight long-term trends in labeling coverage.}
    \label{fig:filter-date-coverage}
\end{figure}

Each anomaly is identified by a unique identifier and may be described by one or more filters.
A single filter can match multiple packets and multiple filters can be associated with the same anomaly.
The dataset is organized in a per-day structure with one capture and its corresponding annotations per calendar day.
Figure~\ref{fig:filter-date-coverage} shows the number of annotations that could be captured while processing the raw data.
A declining trend in coverage is visible in recent years.
As reported by the dataset maintainers~\cite{fontugne2010mawilab}, data collection was stopped in December 2024 due to declining recording quality.

The MAWILab labeling process is implemented as a four-stage pipeline involving an ensemble of unsupervised anomaly detectors, graph-based similarity estimation, clustering and classification.
The procedure is outlined in detail by \cite{fontugne2010mawilab}.
In the first stage, flow-level traffic representations are extracted from the raw packet captures.
In the second stage, these flows are passed to four detectors (\gls{pca}, Gamma, Hough, and \gls{kl}-divergence) with three parameter tunings each.
These detectors operate independently and generate alarms, each representing a segment of suspicious traffic.
The third stage constructs a similarity graph \( G = (V, E) \), where each node \( v_i \in V \) represents an alarm and each edge \( e_{ij} \in E \) quantifies traffic overlap between two alarms using the Simpson index~\cite{fontugne2010mawilab}:
\[
S(T_i,T_j) = \frac{|T_i \cap T_j|}{\min(|T_i|, |T_j|)},
\]
where \( T_i \) and \( T_j \) denote the sets of flows matched by alarms \( v_i \) and \( v_j \), respectively.
Alarms with substantial traffic overlap are connected by edges with non-zero weight.
The Louvain algorithm is applied to this graph to identify communities-groups of alarms that likely describe the same underlying anomaly.

In the fourth stage, each alarm community is classified using SCANN~\cite{merz1999UsingCorrespondenceAnalysis}, a statistical combiner based on correspondence analysis.
A distance score is defined as
\[
d_c = \frac{d_\text{rej}}{d_\text{acc}} - 1,
\]
where \( d_\text{rej} \) and \( d_\text{acc} \) are the distances of the community to the reference point of rejected and accepted communities~\cite{fontugne2010mawilab}.

The final label is determined based on \( d_c \) as follows:
\begin{itemize}
  \item \textbf{anomalous}, if accepted by SCANN.
  \item \textbf{suspicious}, if rejected and \( d_c \le 0 \).
  \item \textbf{notice}, if rejected and \( d_c > 0.5 \).
\end{itemize}

Traffic not covered by any alarm is considered \textbf{benign} by construction and is excluded from the labeled output.
Each labeled anomaly is further annotated with metadata that supports interpretability and downstream filtering:
\begin{itemize}
  \item \textbf{distance}: the SCANN distance score \( d_c \), indicating classification confidence.
  \item \textbf{nbDetectors}: the number of detector configurations that contributed alarms to the community.
  \item \textbf{taxonomy}: a high-level anomaly class (e.g., \texttt{ptmpHTTP}, \texttt{alphfl}, \texttt{ntscICec}) based on observed communication patterns.
  \item \textbf{heuristic}: a numeric tag derived from simple rule-based matching (e.g., 503 for HTTP anomalies, 999 for unknown).
\end{itemize}

In addition, association rule mining is applied to each anomaly's traffic to extract frequent patterns in IP addresses, ports and flow properties.
These patterns act as distinctive identifiers for the anomaly, enabling further classification or blacklisting and are retained as described filters for future reference.

\subsection{Preprocessing Pipeline}\label{subsec:preprocessing-pipeline}

\begin{figure}[ht]
    \centering
    \includegraphics[width=\textwidth]{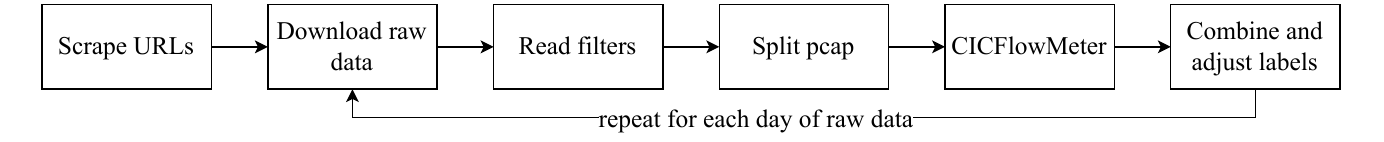}
    \caption{DVC-powered MAWIFlow pipeline converting raw MAWILab captures into labeled flow features.}
    \label{fig:pipeline}
\end{figure}

This paper proposes MAWIFlow, a reproducible pipeline that converts raw MAWILab~v1.1 traffic captures into flow‐based feature sets suitable for \gls{ml} benchmarks.
A flowchart illustrating the general structure of this pipeline is presented in Figure~\ref{fig:pipeline}.

The MAWILab~v1.1 dataset comprises compressed \glspl{pcap} files accompanied by metadata organized as \gls{csv} and \gls{admd} on a daily basis. To ensure data consistency and provenance, \gls{dvc} was employed.
All URLs pointing to compressed packet dumps (\texttt{.pcap.gz}, \texttt{.dump.gz}) and metadata files (\texttt{.csv}, \texttt{.xml}) were extracted via a custom web scraper against the MAWILab website.
Each file reference was registered in a DVC repository and organized in a hive‐partitioned directory hierarchy, enabling selective retrieval.
A \gls{dvc} stage was defined to run per day of raw data.
It ingests all \gls{csv} and \gls{admd} files within a partition, merges filter specifications, and outputs a unified Apache Parquet file.

The next stage executes the following steps for each daily partition:
\begin{enumerate}
  \item Decompression of raw archives.
  \item Packet splitting via the for this purpose developed \texttt{pcap-filter} tool v0.1.0: packets matching each anomaly filter are written to separate \gls{pcap} files and non-matching packets form the benign capture.
  \item Extraction of packet‐level metadata and combination with annotation features. These are stored as byproduct.
  \item Flow generation within a Podman container running CICFlowMeter v3-0.0.4.
  \item Label propagation: flow records inherit MAWILab labels and unlabeled flows are assigned \texttt{benign}.
  \item Aggregation: all flows are concatenated, sorted by timestamp, and written to Apache Parquet file.
\end{enumerate}

To benchmark the baseline \gls{ad} performance, three subsets of approximately $3\times10^{6}$ flows were sampled for January 2011, January 2016 and January 2021.
The selection of these timeframes should ensure temporal coherence within a sample as well as sufficient spacing between samples.
The sample size is chosen according to the size of CIC-IDS-2017.
Numerical features excluding \texttt{Flow ID}, \texttt{Src IP}, \texttt{Src Port}, \texttt{Dst IP}, \texttt{Dst Port}, \texttt{Timestamp} and \texttt{Protocol} were normalized via \texttt{MinMaxScaler} from \texttt{scikit-learn}.
Fitted scalers were archived for reproducibility.
The categorical \texttt{Protocol} feature was encoded using \texttt{OneHotEncoder}.
Labels were binarized: \texttt{benign} $\mapsto0$, all others $\mapsto1$.
Records containing missing values were discarded.

The pipeline ensures reproducible transformation of MAWILab~v1.1 raw data into benchmark‐compatible flow datasets.
The full implementation, including DVC stage definitions and container specifications, is publicly available at \href{https://github.com/TheLurps/MAWIFlow}{GitHub}.

\subsection{Models}\label{subsec:models}

To establish reliable benchmark evaluations and performance baselines, this study utilizes a selection of representative traditional \gls{ml} algorithms along with a \gls{dl}-based architecture.
The traditional \gls{ml} models are implemented with specific configurations as follows:

\begin{itemize}
    \item \textbf{\Glsxtrfull{lr}:}
    Employed as a linear baseline classifier, logistic regression utilizes an optimization strategy based on maximum likelihood estimation.
    Training is performed using the L2 regularization penalty (\texttt{penalty = 'l2'}), an inverse regularization strength parameter (\texttt{C = 1.0}), and a maximum iteration limit increased to 1\,000 to ensure convergence.
    \item \textbf{\Glsxtrfull{dt}:}
    Utilized due to its interpretability and simplicity, the implemented model employs Gini impurity as its splitting criterion.
    Default \texttt{scikit-learn} hyperparameters are applied.
    \item \textbf{\Glsxtrfull{rf}:}
    Deployed as an ensemble improvement over \glspl{dt}, the \gls{rf} uses 100 estimators (\texttt{n\_estimators = 100}), a maximum number of sampled features per split being the square root of the total features (\texttt{max\_features = 'sqrt'}), and entropy-based splitting for robustness and generalization.
    \item \textbf{XGBoost:}
    Selected for its demonstrated predictive accuracy on structured tabular data, the implemented configuration includes 100 boosting rounds (\texttt{n\_estimators = 100}), a learning rate fixed at 0.1, maximum depth restricted to 6 to avoid overfitting, and tree subsampling parameters (\texttt{subsample = 0.8; colsample\_bytree = 0.8}) introduced to reduce variance and enhance generalization properties.
\end{itemize}

The deep learning baseline involves a hybrid neural network architecture combining convolutional layers and recurrent units for robust spatio-temporal feature extraction. The \gls{cnnbilstm} proposed by \cite{wang2024new} was modified from a multi-class classifier to binary:

\begin{itemize}
    \item \textbf{\gls{cnnbilstm}:}
    This architecture sequentially integrates one-dimensional convolution layers with \gls{bilstm} layers.
    Specifically, the neural network architecture is defined as follows:
    \begin{itemize}
        \item An initial \texttt{Conv1D} layer applies 64 convolutional filters, followed by a ReLU activation, \texttt{MaxPooling1D} (\texttt{pool\_size = 8}), and batch normalization.
        \item A \gls{bilstm} layer subsequently processes the feature maps extracted by the convolutional stage, operating with 32 units per direction, yielding a combined 64-dimensional feature vector.
        \item A second pooling step employing \texttt{MaxPooling1D} (\texttt{pool\_size = 16}), enhanced by batch normalization, further compresses the temporal representation.
        \item Another \gls{bilstm} layer with 64 units per direction processes this compressed representation, followed by a dropout layer (\texttt{dropout\_rate = 0.6}), introduced to reduce overfitting.
        \item A final dense layer with sigmoid activation generates the binary classification probability.
    \end{itemize}
\end{itemize}

Frameworks and package versions supporting each model's reproducibility include scikit-learn v1.6.1 for \gls{lr}, \gls{dt} and \gls{rf}; XGBoost v3.0.2 for gradient boosting models; and TensorFlow v2.19.0 with Keras API for the \gls{cnnbilstm} architecture.

\subsection{Training pipeline}\label{subsec:training-pipeline}

To ensure reproducibility and consistent evaluation, this study defined a structured training and testing pipeline applied uniformly across all employed models.
After performing data preprocessing steps (Section~\ref{subsec:preprocessing-pipeline}) datasets for traditional \gls{ml} methods were split into training and testing subsets, following an 80\%/20\% ratio.
Evaluated datasets include the CIC-IDS-2017 benchmark and the MAWI\-Lab subsets (Section~\ref{subsec:mawilab-dataset}).
For the \gls{cnnbilstm} model, an additional validation subset was created: initially, an 80\%/20\% training–test dataset split was performed, followed by a secondary 80\%/20\% partitioning of the training subset to yield distinct training and validation sets.
\gls{cnnbilstm} training proceeded for up to 50 epochs using batches of 32 samples.
Optimization employed the Adam optimizer (\texttt{learning\_rate = 0.001}) and gradient norm clipping (\texttt{clipnorm = 1.0}) to maintain numerical stability.
Early stopping (monitoring validation loss with a patience parameter of five epochs) was implemented to prevent overfitting when training on each individual dataset, with model weights restored from the epoch achieving minimal validation loss.

For experiments involving temporally combined datasets, early stopping was explicitly disabled to facilitate comprehensive learning across heterogeneous data distributions.
Under these combined training conditions, \gls{cnnbilstm} models consistently trained for the full 50 epochs.

\section{Evaluation}\label{sec:evaluation}

The evaluation compares five baseline models (Section~\ref{subsec:models}) across multiple datasets.
Specifically considered datasets comprise the CIC-IDS-2017 benchmark and temporally distinct subsets of MAWILab~v1.1 sampled from January of 2011, 2016, and 2021 using MAWIFlow.
Each dataset was divided into a training, validation and test subset.
Performance was assessed through accuracy, recall, precision, F1-score, and \gls{auc} on the test subset when the training subset was used otherwise the whole dataset was evaluated.
Table~\ref{tab:baseline-results} summarizes the results of these experiments.
Additional cross-temporal evaluations examined model robustness using distinct training and testing periods or their combinations.
Figure~\ref{fig:roc-curves} visualizes these cross-temporal experiments.

\subsection{Baseline Model Performance Results}\label{subsec:baseline-model-performance-results}

\begin{table}[ht]
\centering
\small
\caption{Classifier performance (Accuracy, Recall, Precision, F1-score, \gls{auc}) on CIC-IDS-2017 and MAWILab~v1.1 subsets (January 2011, 2016, 2021).}
\label{tab:baseline-results}
\begin{tabular}{llccccc}
\toprule
Dataset & Model & Accuracy & Recall & Precision & F1 & \gls{auc} \\
\midrule
\multirow{5}{*}{\shortstack[l]{CIC-IDS-2017}} 
    & \glsxtrshort{cnnbilstm}             & 0.9842   & 0.9647  & 0.9559    & 0.9603   & 0.9985  \\
    & \Glsxtrlong{dt} & \textbf{0.9990} & \textbf{0.9983} & 0.9965    & \textbf{0.9974} & 0.9990  \\
    & \Glsxtrlong{lr}     & 0.9257   & 0.7679  & 0.8424    & 0.8034   & 0.9718  \\
    & \Glsxtrlong{rf} & 0.9989   & 0.9978  & \textbf{0.9966} & 0.9972   & \textbf{0.9999} \\
    & XGBoost          & 0.9986   & 0.9971  & 0.9961    & 0.9966   & \textbf{0.9999} \\
\midrule
\multirow{5}{*}{\shortstack[l]{MAWILab~v1.1\\2011-01}} 
    & \glsxtrshort{cnnbilstm}             & 0.8460   & 0.9147  & 0.8762    & 0.8950   & 0.9014  \\
    & \Glsxtrlong{dt} & 0.8334   & 0.8822  & 0.8853    & 0.8837   & 0.7955  \\
    & \Glsxtrlong{lr}     & 0.7899   & \textbf{0.9361} & 0.8036    & 0.8648   & 0.8178  \\
    & \Glsxtrlong{rf} & \textbf{0.8579} & 0.9131  & \textbf{0.8916} & \textbf{0.9022} & 0.9060  \\
    & XGBoost          & 0.8535   & 0.9269  & 0.8761    & 0.9008   & \textbf{0.9097} \\
\midrule
\multirow{5}{*}{\shortstack[l]{MAWILab~v1.1\\2016-01}}
    & \glsxtrshort{cnnbilstm}             & 0.7843   & 0.7764  & 0.8291    & 0.8019   & 0.8458  \\
    & \Glsxtrlong{dt} & 0.8237   & 0.8614  & 0.8312    & 0.8460   & 0.8507  \\
    & \Glsxtrlong{lr}     & 0.7504   & 0.7202  & 0.8145    & 0.7644   & 0.8100  \\
    & \Glsxtrlong{rf} & \textbf{0.8321} & \textbf{0.8869} & 0.8271    & \textbf{0.8559} & \textbf{0.9023} \\
    & XGBoost          & 0.7930   & 0.7799  & \textbf{0.8404} & 0.8091   & 0.8565  \\
\midrule
\multirow{5}{*}{\shortstack[l]{MAWILab~v1.1\\2021-01}} 
    & \glsxtrshort{cnnbilstm}             & 0.8329   & 0.8900  & 0.8849    & 0.8874   & 0.8915  \\
    & \Glsxtrlong{dt} & 0.8199   & 0.8771  & 0.8792    & 0.8782   & 0.7733  \\
    & \Glsxtrlong{lr}     & 0.7798   & \textbf{0.9155} & 0.8113    & 0.8602   & 0.8046  \\
    & \Glsxtrlong{rf} & \textbf{0.8407} & 0.8993  & \textbf{0.8870} & \textbf{0.8931} & \textbf{0.9034} \\
    & XGBoost          & 0.8361   & 0.8956  & 0.8844    & 0.8900   & 0.8951  \\
\bottomrule
\end{tabular}
\end{table}

On the CIC-IDS-2017 dataset, \gls{dt}, \gls{rf} and XGBoost classifiers achieved nearly perfect scores, with accuracy values of above 0.9986 and \gls{auc} scores of above 0.9990.
\gls{cnnbilstm} showed slightly lower but still strong performance across all metrics, while \gls{lr} yielded visibly weaker results.

For the MAWILab January 2011 subset, all classifiers demonstrated decreased performance in comparison to CIC-IDS-2017. \gls{rf} remained best performing classifier, closely followed by \gls{cnnbilstm}, while \gls{dt} and XGBoost yielded moderate results.
\gls{lr} demonstrated the poorest performance among the models assessed, aside from having the highest recall.

In the MAWILab January 2016 subset, classifier performances decreased further relative to CIC-IDS-2017 and the earlier 2011 subset.
\gls{rf} remained superior, closely followed by \gls{dt}. \gls{cnnbilstm} exhibited notably reduced accuracy, precision, and recall, while \gls{lr} continued to display clearly weaker performance compared to other classifiers.

Finally, on the MAWILab January 2021 subset, model performance partially rebounded compared to the 2016 subset.
\gls{rf} again achieved the highest performance metrics, with \gls{cnnbilstm} performing nearly as strongly.
\gls{dt} and XGBoost showed intermediate results, and \gls{lr} preserved its position as the lowest-performing classifier except for its recall.

In summary, the evaluated classifiers exhibited highly dataset-dependent performances. On CIC-IDS-2017, \gls{rf}, \gls{dt} and XGBoost delivered nearly ideal results, all achieving accuracy scores close to 1.0 and \gls{auc} scores nearing perfection.
Conversely, on the MAWILab subsets, all models presented notably lower performance metrics compared to CIC-IDS-2017 data.
\gls{rf} consistently emerged as the top classifier across the MAWILab subsets, with \gls{dt} and \gls{cnnbilstm} typically providing intermediate but varying results.
\gls{lr} consistently demonstrated the weakest performance on every analyzed dataset.

\subsection{Cross-Temporal Generalization}\label{cross-temporal-generalization}

\begin{figure}[!ht]
  \centering

  \begin{subfigure}[b]{0.48\textwidth}
    \centering
    \includegraphics[width=0.8\textwidth]{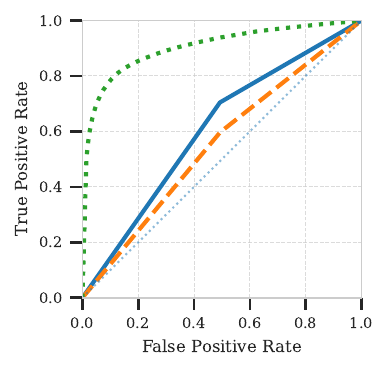}
    \caption{Random Forests}
    \label{fig:roc-curves-rf}
  \end{subfigure}
  \hfill
  \begin{subfigure}[b]{0.48\textwidth}
    \centering
    \includegraphics[width=0.8\textwidth]{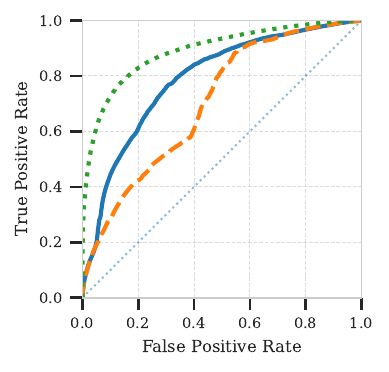}
    \caption{\glsxtrshort{cnnbilstm}}
    \label{fig:roc-curves-bilstm-single}
  \end{subfigure}

  \vspace{1em}

  \begin{subfigure}[b]{0.48\textwidth}
    \centering
    \includegraphics[width=0.8\textwidth]{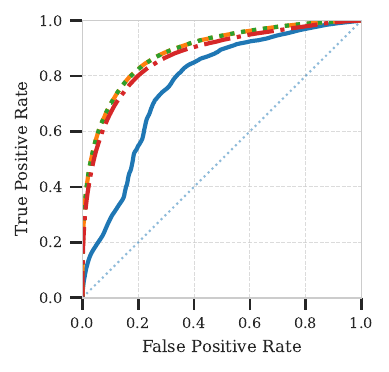}
    \caption{\glsxtrshort{cnnbilstm} (Combined Datasets)}
    \label{fig:roc-curves-bilstm-multiple}
  \end{subfigure}
  \hfill
  \begin{subfigure}[b]{0.48\textwidth}
    \centering
    \small
    \begin{adjustbox}{max width=\textwidth}
    \begin{tabular}{lllccc}
      \toprule
      Model & Training set & Style & Acc & F1 & \gls{auc} \\
      \midrule
      Random Forest & 2011-01 & \linestylebox{solid}{plot1}  & 0.65 & 0.75 & 0.61 \\
                    & 2016-01 & \linestylebox{dashed}{plot2} & 0.57 & 0.67 & 0.55 \\
                    & 2021-01 & \linestylebox{dotted}{plot3} & \textbf{0.84} & \textbf{0.89} & \textbf{0.90} \\
      \midrule
      \glsxtrshort{cnnbilstm}    & 2011-01 & \linestylebox{solid}{plot1}  & 0.78 & 0.86 & 0.79 \\
                    & 2016-01 & \linestylebox{dashed}{plot2} & 0.75 & 0.83 & 0.70 \\
                    & 2021-01 & \linestylebox{dotted}{plot3} & \textbf{0.83} & \textbf{0.89} & \textbf{0.89} \\
      \midrule
      \glsxtrshort{cnnbilstm}    & 2011/16-01      & \linestylebox{solid}{plot1}   & 0.79 & 0.86 & 0.77 \\
                    & 2011/21-01      & \linestylebox{dashed}{plot2}  & \textbf{0.84} & \textbf{0.89} & \textbf{0.89} \\
                    & 2016/21-01      & \linestylebox{dotted}{plot3}  & 0.84 & 0.89 & 0.89 \\
                    & 2011/16/21-01   & \linestylebox{dash dot}{plot4}& 0.81 & 0.86 & 0.87 \\
      \bottomrule
    \end{tabular}
    \end{adjustbox}
    \caption{Summary of accuracy, F1-score and AUC.}
    \label{fig:roc-curves-table}
  \end{subfigure}

  \caption{Evaluation of model generalization on the MAWILab~v1.1 January 2021 sample, illustrating how training data selection affects classification performance.}
  \label{fig:roc-curves}
\end{figure}

Figure~\ref{fig:roc-curves} shows the results of \glspl{rf} and \gls{cnnbilstm} models trained on different samples and combinations of data processed via MAWIFlow.
All models were tested on the 2021 subset; if the 2021 data was used during training, the models have been tested on the test subset thereof.
Additionally to the \gls{roc} curve, accuracy, F1-score and \gls{auc} are displayed in Table~\ref{fig:roc-curves-table}.

As shown in Section~\ref{subsec:baseline-model-performance-results}, \glspl{rf} is the best performing model when trained and tested within the same sample.
But as visualized in Figure~\ref{fig:roc-curves-rf}, training on older data results in drastic performance degradation.
The model's \gls{auc} metric declines to 0.61 and 0.55 when trained on the 2011 and 2021 datasets, respectively, indicating that the model performs at a near-random level.

Figure~\ref{fig:roc-curves-bilstm-single} depicts the \gls{roc} curves of \gls{cnnbilstm} models trained on the same samples as the \glspl{rf}.
Here, the performance drop is less dominant, although its initial performance when trained on the 2021 training subset is worse.
To quantify the \gls{rf} drops from 0.90 \gls{auc} by 0.29 to 0.61 when trained on 2011, where the \gls{cnnbilstm} only drops from 0.89 \gls{auc} by 0.10 to 0.79.
The reduction in performance is also evident in accuracy and F1-score.

Training the \gls{cnnbilstm} on a sequential concatenation of samples results in prediction performances as shown in Figure~\ref{fig:roc-curves-bilstm-multiple}.
It can be observed that combining samples 2011 and 2016 increases accuracy slightly, while keeping F1-score consistent and decreasing \gls{auc} in comparison to the model training on only 2011.
The inclusion of the 2021 training subset boosts performance marginally above the metrics of the \gls{cnnbilstm} solely trained on the 2021 subset.
A decline in performance is evident when all three samples are combined for training.

Cross-temporal evaluation reveals significant performance degradation when classifiers trained exclusively on older dataset samples are tested on more recent subsets.
Specifically, the \gls{rf} classifier exhibited unsuitable performance under temporal drift conditions.
By contrast, \gls{cnnbilstm} consistently achieved slightly better generalization.

\FloatBarrier

\section{Discussion}\label{sec:discussion}

The evaluation presented in this study yields three primary insights regarding synthetic benchmark datasets and the temporal robustness of selected anomaly detection methods for network traffic.

Firstly, experimental results on CIC-IDS-2017 indicate that current \gls{ml} methods achieve consistently high performance metrics.
This suggests limited suitability of this dataset for realistic benchmark purposes.
Specifically, classifiers such as \gls{rf} and \gls{cnnbilstm} obtained nearly perfect classification scores (\gls{auc} > 0.99) when trained and tested on CIC-IDS-2017 data.
This near-perfect separability aligns with recent findings by \cite{cantone2024CrossDatasetGeneralizationMachine}, who demonstrate similarly ideal within-dataset performance for fixed IDS benchmarks, which sharply degrades under cross-dataset evaluation. Thus, while CIC-IDS-2017 has been broadly adopted, its low difficulty provides an overly optimistic picture of the models' real-world performance and limits its value as a contemporary evaluation standard.

Secondly, the evaluation demonstrates that traditional tree-based models, particularly \gls{rf}, yield strong classification performance within temporally static test conditions.
For example, \gls{rf} classifiers trained and evaluated on identical time periods consistently achieved high discrimination metrics, with an \gls{auc} of approximately 0.91 on the January 2011 sample.

Thirdly, cross-temporal experiments revealed a clear decline in the performance of the \gls{rf} classifier when trained on older data and tested on more recent samples.
Notably, the performance of a \gls{rf} trained on the January 2011 subset dramatically declined to an \gls{auc} of 0.61 when directly evaluated on traffic data from January 2021.
In contrast, \gls{cnnbilstm} models showed improved temporal robustness under identical conditions, achieving an \gls{auc} of approximately 0.79 on the 2021 sample when trained exclusively on the older 2011 data.
Such hybrid architectures like the \gls{cnnbilstm} have previously been shown by \cite{duraj2025DetectionAnomaliesData} to effectively learn persistent sequential patterns as well as dynamic temporal relationships in network data streams, reducing their susceptibility to overfitting to historical distribution contexts.

Nevertheless, several methodological limitations must be explicitly acknowledged.
First, the anomaly labels in MAWIFlow---derived from the MAWILab labeling pipeline---reflect outcomes from heuristic-based ensemble anomaly detection methods, graph-based clustering algorithms and statistical combiners. 
Therefore, the derived labels inherently contain uncertainty and potential biases, potentially distorting classification metrics regarding true anomaly detection performance.
Additionally, the selection of data exclusively from January 2011, 2016 and 2021 imposes a temporal sampling limitation.

\section{Conclusion}\label{sec:conclusion}

In conclusion, this research critically assessed various \gls{ad} techniques, focusing on both traditional \gls{ml} classifiers and the \gls{cnnbilstm} model, using both an established benchmark and real-world network datasets.
The analysis revealed that the CIC-IDS-2017 dataset's near-perfect separability limits its effectiveness as an evaluation tool.
Therefore, a shift toward benchmarks such as MAWIFlow, which evaluate temporal shifts, is necessary to better reflect real-world deployment conditions.

Within the study's temporal samples, it was observed that \gls{rf} models demonstrated strong \gls{ad} capabilities, often surpassing \gls{cnnbilstm} in raw discrimination metrics.
However, these strengths diminished severely across cross-temporal evaluations, indicating a deterioration in models trained on outdated data, which significantly limits their practical applicability.
On the contrary, \gls{cnnbilstm} models showed enhanced robustness in comparision to \gls{rf} when trained on outdated data.

Furthermore, empirical analysis suggested that merging temporally adjacent training subsets modestly improved model generalization, demonstrating the benefit of targeted temporal aggregation.
Conversely, aggregating data spanning too broad a temporal range diluted predictive accuracy due to conflicting dynamics and distributional shifts.

Future research should address current limitations.
Network attacks are complex and continuously evolving, suggesting a need for expanded analyses incorporating packet-level payloads and fine-grained traffic measurements.
Additionally, adopting adaptive training paradigms, such as incremental training or ensemble drift-adaptation, could potentially enhance model robustness.
Finally, exploring transferability and cross-domain generalization by training on one network environment and testing on entirely different settings could yield insights into developing universally effective \gls{ad} frameworks.
Evaluation could also benefit from a broader spectrum of models, including time-aware ensemble methods and unsupervised detectors, alongside systematic hyperparameter tuning strategies, such as cross-validation or Bayesian optimization.
Multiclass classification techniques should be explored to address real-world scenarios more effectively, enhancing the development of resilient \gls{ids} models capable of adapting to dynamic network environments.

\section*{Acknowledgments}
The authors employed GPT-o3 and GPT-4.5 by OpenAI for improvement of spelling and grammar refinement across all sections of this manuscript. Every AI-assisted passage was independently verified, edited, and remains the sole responsibility of the authors.

\bibliographystyle{unsrt}  
\bibliography{references}  

\end{document}